\documentclass[supp]{new_tlp} 
\usepackage{aopmath}

\usepackage{pifont,graphicx,latexsym}
\usepackage{xspace,url}
\usepackage[T1]{fontenc}
\usepackage[breaklinks,hidelinks]{hyperref}

\newtheorem{example}{Example}[section]

\newcommand\tabdual{\textsc{Tabdual}\xspace}
\newcommand\er{\textsc{Evolp/r}\xspace}

\newcommand{\til}{\raise.17ex\hbox{$\scriptstyle\mathtt{\sim}$}}

\newcommand\bcmdtab{\noindent\bgroup\tabcolsep=0pt%
  \begin{tabular}{@{}p{10pc}@{}p{20pc}@{}}}
\newcommand\ecmdtab{\end{tabular}\egroup}

\usepackage{supp}

\submitted{n/a}
\revised{n/a}
\accepted{n/a}

\begin{document}

  \title[Joint Tabling of Logic Program Abductions and Updates]
        {Joint Tabling of\\ Logic Program Abductions and Updates}

  \author[A. Saptawijaya and L. M. Pereira]
         {ARI SAPTAWIJAYA$^{\footnotesize \thanks{Affiliated with Fakultas Ilmu Komputer at Universitas Indonesia, Depok, Indonesia.}}$ and LU\'IS MONIZ PEREIRA\\
                  Centro de Intelig\^encia Artificial (CENTRIA)\\
                   Departamento de Inform\'atica, Faculdade de Ci\^encias e Tecnologia\\ 
Universidade Nova de Lisboa, 2829-516 Caparica, Portugal\\
         \email{ar.saptawijaya@campus.fct.unl.pt, lmp@fct.unl.pt}}

\pagerange{\pageref{firstpage}--\pageref{lastpage}}
\volume{\textbf{10} (3)}
\jdate{March 2002}
\setcounter{page}{1}
\pubyear{2002}

\pubauthor{Saptawijaya}
\jurl{xxxxxx}
\pubdate{22 June 2013}

\maketitle

\label{firstpage}

\begin{abstract}
Abductive logic programs offer a formalism to declaratively represent and reason about problems  in a variety of areas: diagnosis, decision making, hypothetical reasoning, etc. On the other hand, logic program updates allow us to express knowledge changes, be they internal (or self) and external (or world) changes. Abductive logic programs and logic program updates thus naturally coexist in problems that are susceptible to hypothetical reasoning about change. Taking this as a motivation, in this paper we integrate abductive logic programs and logic program updates by jointly exploiting tabling features of logic programming. The integration is based on and benefits from the two implementation techniques we separately devised previously, viz., tabled abduction and incremental tabling for query-driven propagation of logic program updates. A prototype of the integrated system is implemented in XSB Prolog.
\end{abstract}


\section{Introduction}
Abduction has been well studied in logic programming \cite{ds92-sldnfa,is96-fixpoint,fk97-iff,egl97-abduction,kkt98-alp,si00-abductionTMS,aps04-abdual}, and it offers a formalism to declaratively represent and reason about problems  in a variety of areas. Furthermore, the progress of logic programming promotes new techniques for implementing abduction in logic programs. For instance, we have shown recently in \cite{iclp13-tabdual}, that abduction may benefit from tabling mechanisms; the latter mechanisms are now supported by a number of Prolog systems, to different extent. In that work, tabling is employed to reuse priorly obtained abductive solutions from one abductive context to another, thus avoiding potential unnecessary recomputation of those solutions.

Given the advances of tabling features, like incremental tabling \cite{saha06-incremental} and answer subsumption \cite{sw10-answer}, we have also explored these in addressing logic program updates. Our first attempt, reported in \cite{lpnmr13-evolpr}, exploits incremental tabling of fluents in order to automatically maintain the consistency of program states, analogously to assumption based truth-maintenance system, due to assertion and retraction of fluents. Additionally,  answer subsumption of fluents allows to address the frame problem by automatically keeping track, at low level, of their latest assertion or retraction, whether as a result of updated facts or concluded by rules. In \cite{lpar19-evolpr}, the approach is improved, by fostering further incremental tabling. It leaves out the superfluous use of the answer subsumption feature, but nevertheless still allows direct access to the latest time a fluent is true, via system table inspection predicates. In the latter approach, incremental assertions of fluents automatically trigger \emph{system level} incremental upwards propagation and tabling of fluent updates, on the initiative of top goal queries (i.e., by need only). The approach affords us a form of controlled (i.e., query-driven) but automatic truth-maintenance (i.e., automatic updates propagation via incremental tabling), up to actual query time.

When logic programs are used to represent agent's knowledge, then the issue of logic program updates pertains to expressing knowledge updates. Many applications of abduction, as in reasoning of rational agents and decision making, are typically susceptible to knowledge updates and changes, whether or not hypothetical. Thus, abductive logic programs and logic program updates naturally coexist in these applications. Taking such applications as a motivation, one of which we currently pursue \cite{padl14-moral}, here we propose an implementation approach to integrate abductive logic programs and logic program updates by exploiting together tabling features of logic programming. The integration is strongly based on the reported approaches implemented in our two systems: \tabdual \cite{iclp13-tabdual} for tabled abduction, and \er \cite{lpar19-evolpr} for query-driven propagation of logic program updates with incremental tabling. In essence, we show how tabled abduction is jointly combined with incremental tabling of fluents in order to benefit from each feature, i.e., abductive solutions can be reused from one context to another, while also allowing query-driven, system level, incremental fluent update upwards propagation. The integration is achieved by a program transformation plus a library of reserved predicates. The different purposes of the dual program transformation, employed both in \tabdual and \er, are now consolidated in one integrated program transformation: on the one hand, it helps to efficiently deal with downwards by-need abduction under negated goals; on the other hand, it helps to incrementally propagate upwards the dual negation complement of a fluent.

The paper is organized as follows. Section \ref{sec:prelim} recaps tabled abduction and logic program updates with incremental tabling. We detail our approach to the integration in Section \ref{sec:integrate}, and conclude, in Section \ref{sec:conclude}, by mentioning related and future work.

\section{\tabdual and \er}\label{sec:prelim}

\paragraph{{\bf Tabled Abduction (\tabdual)}} We illustrate the idea of tabled abduction. Consider an abductive logic program $P_0$, with $a$ and $b$ abducibles: 

{\centering

$q \leftarrow a$. \qquad $s \leftarrow b, q$. \qquad $t \leftarrow s, q$.\\
}
\noindent
Suppose three queries: $q$, $s$, and $t$, are individually launched, in that order. The first query, $q$, is satisfied simply by taking $[a]$ as the abductive solution for $q$, and tabling it. Executing the second query, $s$, amounts to satisfying the two subgoals in its body, i.e., abducing $b$ followed by invoking $q$. Since $q$ has previously been invoked, we can benefit from reusing its solution, instead of recomputing, given that the solution was tabled. I.e., query $s$ can be solved by extending the current ongoing abductive context $[b]$ of subgoal $q$ with the already tabled abductive solution $[a]$ of $q$, yielding $[a,b]$. The final query $t$ can be solved similarly. Invoking the first subgoal $s$ results in the priorly registered abductive solution $[a,b]$, which becomes the current abductive context of the second subgoal $q$. Since $[a,b]$ subsumes the previously obtained (and tabled) abductive solution $[a]$ of $q$, we can then safely take $[a,b]$ as the abductive solution to query $t$. This example shows how $[a]$, the abductive solution of the first query $q$, can be reused from one abductive context of $q$ (i.e., $[b]$ in the second query, $s$) to its other context (i.e., $[a,b]$ in the third query, $t$). In practice the body of rule $q$ may contain a huge number of subgoals, causing potentially expensive  recomputation of its abductive solutions, if they are not tabled. 

Tabled abduction with its prototype \tabdual,  implemented in XSB Prolog \cite{sw12-xsb}, consists of a program transformation from abductive normal logic programs into tabled logic programs; the latter are self-sufficient program transforms, which can be directly run to enact abduction by means of \tabdual's library of reserved predicates. We recap the key points of the transformation. First, for every predicate $p$ with arity $n$ ($p/n$ for short) defined in a program, two new predicates are introduced in the transform:  $p_{ab}/(n+1)$ that tables one abductive solution for $p$ in its single extra argument, and $p/(n+2)$ that reuses the tabled solution of $p_{ab}$ to produce a solution from a given input abductive context into an output abductive context (both abductive contexts are the two extra arguments of $p$). The role of  abductive contexts is important, e.g., in contextual abductive reasoning, cf. \cite{pdh14-contextual}. Second, for abducing under negative goals, the program transformation employs the \emph{dual transformation} \cite{aps04-abdual}, which makes negative goals `positive' literals, thus permitting to avoid the computation of all abductive solutions of the positive goal argument, and then having to negate their disjunction. The dual transformation enables us  to obtain one abductive solution at a time, just as when we treat abduction under positive goals. In essence, the dual transformation defines for each atom $A$ and its set of rules $R$ in a normal program $P$, a set of dual rules whose head $not\_A$ is true if and only if $A$ is false by $R$ in the considered semantics of $P$. Note that, instead of having a negative goal $not\ A$ as the rules' head, we use its corresponding `positive' literal, $not\_A$.  The reader is referred to \cite{iclp13-tabdual} and publications cited thereof for detailed aspects of tabled abduction.

\paragraph{{\bf Logic Program Updates with Incremental Tabling (\er)}}
\er follows the paradigm of Evolving Logic Programs (EVOLP) \cite{ablp02-evolp}, by adapting its syntax and semantics, but simplifies it by restricting updates to fluents only. Syntactically, every fluent $F$ is accompanied by its fluent complement $\til F$. Program updates are enacted by having the reserved predicate $assert/1$ in the head of a rule, which updates the program by fluent $F$, whenever the assertion $assert(F)$ is true in a model; or retracts $F$ in case $assert(\til F)$ obtains in the model under consideration. Though updates in \er are restricted to fluents only, it nevertheless still permits rule updates by introducing a rule name fluent that uniquely identifies the rule for which it is introduced. Such a rule name fluent is placed in the body of a rule to turn the rule on and off, cf. \cite{poole88-default}; this being achieved by asserting or retracting that specific fluent. The reader is referred to \cite{lpar19-evolpr} for a more detailed theoretical basis of \er. 

Like \tabdual, \er is implemented by a compiled program transformation plus a library of reserved predicates. The implementation makes use of incremental tabling \cite{saha06-incremental}, a feature in XSB Prolog that ensures the consistency of answers in a table with all dynamic clauses on which the table depends by incrementally maintaining the table, rather than by  recomputing answers in the table from scratch to keep it updated. The main idea of the implementation is described as follows. The input program is first transformed and then the initialization phase takes place. It sets a predefined upper global time limit in order to avoid potential iterative non-termination of updates propagation and it additionally creates and initializes the table for every fluent. When fluent updates are given, they are initially kept pending in the database, and only on the initiative of top-goal queries, i.e., by need, incremental assertions make these pending updates become active (if not already so), but only those with timestamps up to an actual query time. Such assertions automatically trigger system-implemented incremental upwards propagation of updates and tabling of fluents (thanks to the incremental tabling). Because fluents are tabled, a direct access to the latest time a fluent is true can be made possible by means of existing table inspection predicates, and thus recursion through the frame axiom can be avoided. Consequently, in order to establish whether a fluent $F$ is true at an actual query time, it suffices to inspect in the table the latest time both $F$ and its complement $\til F$ are true, and to verify whether $F$ is supervened by $\til F$.

We recap the key points of the transformation. First, the transformation adds to each program clause of fluent $f/n$ the timestamp information that figures as the only extra argument of fluents (i.e., heads of clauses) and denotes a point in time when a fluent is true (known as \emph{holds-time}). Having this extra argument, both fluent $f/(n+1)$ and its complement $\til f/(n+1)$ are declared as dynamic and incremental. Second, each fluent (goal) $G$ in the body of a clause is called via a reserved \emph{incrementally} tabled predicate $fluent(G,H_G)$ that non-deterministically returns holds-time $H_G$ of fluent $G$. In essence, this reserved predicate simply calls $G$ and obtains $H_G$ from $G$'s holds-time argument. Since every fluent and its complement are \emph{incrementally} dynamic, the dependency of the incrementally tabled predicate $fluent/2$ on them can be correctly maintained. Third, the holds-time of fluent $f$ in the head of a clause is determined by which \emph{inertial} fluent in its body holds \emph{latest}. Fourth, the dual transformation from \tabdual is adapted for helping propagate the dual negation complement $\til F$ of a fluent $F$ incrementally, making the holds-time of $\til F$ (and other fluents that depend on it) also available in the table. 

\section{Integrating \tabdual and \er}\label{sec:integrate}
When logic programs are used to represent agent's knowledge with abduction for decision making, such applications are typically susceptible to knowledge updates and changes, e.g., because of incomplete and imprecise knowledge, hypothetical updates, and changes caused by agent's actions (side-effects). Driven by such applications, one of which we are currently pursuing \cite{padl14-moral}, and given that  \tabdual and \er have been conceptualized to deal with abduction and logic program updates independently, our subsequent challenge is how to seamlessly integrate both approaches. In Section \ref{sec:prelim} we observe that tabling is employed both in \tabdual and \er, despite its different purposes. Therefore, in addition to enable abduction and knowledge updates in a unified approach, the integration also aims at keeping the different purposes served by tabling in \tabdual and \er. That is, on the one hand the integration should allow reusing an abductive solution entry from an abductive context to another. On the other hand, it should also support system level incremental upwards updates propagation. We now detail an approach to achieve these aims through a program transformation and library of reserved predicates.

\paragraph{{\bf Enabling Abducibles}}\label{sec:enabling}
In abduction it is desirable to generate only abductive explanations relevant for the problem at hand. One stance for selectively enabling the assumption of abducibles in abductive logic programs is introducing rules encoding domain specific information about which particular assumptions are to be considered in a specific situation. We follow the approach proposed in \cite{pdpl13-inspecting}, i.e., the notion of expectation is employed to express preconditions for enabling the assumption of an abducible. An abducible $A$ can be assumed only if there is an expectation for it, and there is no expectation to the contrary. We say then that the abducible is \emph{considered}, expressed by the rule:

{\centering

$consider(A) \leftarrow expect(A),\ not\ expect\_not(A),\ A$.\\
}
\noindent
This method requires program clauses with abducibles to be preprocessed. That is, for every abducible $A$ appearing in the the body of a rule, $A$ is substituted with $consider(A)$. For instance, given abducible $a$, rule $p \leftarrow a$ is preprocessed into rule $p \leftarrow consider(a)$.

\paragraph{{\bf The Roles of Abductive Contexts and Holds-Time}}\label{sec:roles}
In scientific reasoning tasks, it is common that besides the need to abductively discover which hypotheses to assume in order to justify some observation, one may also want to know some of the side-effects of those assumptions. This is one important extension of abduction, viz., to verify whether some secondary observations are plausible in the presence of already obtained abductive explanations, i.e., in the abductive context of the primary one. 

As in \tabdual, our integration makes use of abductive contexts. They permit a mechanism for reusing already obtained abductive solutions, which are tabled, from one context to another. Technically, this is achieved by having two types of abductive context: \emph{input} and \emph{output}, where an abductive solution is in the output context and obtained from the input context plus a tabled abductive solution. In Section \ref{sec:prelim} we show that these two contexts figure as extra arguments of a predicate. 

Updates due to new observations or changes caused by side-effects of abductions may naturally occur, and from the logic program updates viewpoint the time when such changes or updates take place needs to be properly recorded. In \er, this is maintained via the timestamp information, known as holds-time, that figures as an extra argument in a fluent predicate. Like in \er, this timestamp information plays an important role in the integration for propagating updates and tabling fluents affected by these propagations, as shown in subsequent sections.

Based on the need for abductive contexts and holds-time, every predicate $p/n$, i.e.,  $p(X_1,\dots,X_n)$ is now transformed into $p(X_1,\dots,X_n,I,O,H)$, where the three extra arguments refer to the input context $I$, the output context $O$, and the timestamp $H$.

We next show the mechanisms to compute abductive solutions and maintain holds-time through updates propagation using the ingredients discussed earlier. 
\begin{example}\label{ex:incr-tabling}
Consider $P_1$ with abducible $a$:  \qquad\qquad $q \leftarrow a$. \qquad $expect(a)$.
\end{example}
\noindent
After preprocessing abducible $a$ in the body of rule $q\leftarrow a$, cf. ``Enabling Abducibles'', we have the program:

{\centering

$q \leftarrow consider(a)$. \qquad $expect(a)$.\\
}
\noindent
The preprocessed program is now ready to transform. We first follow the rule name fluent mechanism of \er, i.e., a unique rule name fluent of the form $\#r(Head,Body)$ is assigned to each rule $Head \leftarrow Body$. For this example, we have only one rule, i.e., $q\leftarrow consider(a)$, which is assigned the rule name fluent $\#r(q,[consider(a)])$. Recall, the rule name fluent is used to turn the corresponding rule on and off by introducing it in the body of the rule. Thus, we have:

{\centering

$q \leftarrow \#r(q,[consider(a)]), consider(a)$. \qquad\qquad $expect(a)$.\\
}
\noindent
Next, we attach the three additional arguments described earlier. For clarity of explanation, we do that in two steps: first, we add abductive context arguments and discuss how abductive solutions are obtained from them; second, we include the timestamp argument for the purpose of maintaining holds-time in updates propagation.
\paragraph{{\bf Finding Abductive Solutions}}
Adding abductive contexts brings us to the transform below ($cons$ is shorthand for $consider$):

{\centering

$q(I,O) \leftarrow \#r(q,[cons(a)],I,R), cons(a,R,O)$.\qquad\qquad $expect(a,I,I)$.\\
}
\noindent
The abductive solution of $q$ is obtained in its output abductive context $O$ from its input context $I$, by relaying the \emph{ongoing} abductive solution stored in context $R$ from subgoal $\#r(q,[cons(a)],I,R)$ to subgoal $cons(a,R,O)$ in the body. For $expect(a)$, the content of the context $I$ is simply relayed from the input to the output context. That is, having no body, the output context does not depend on the context of any other goals, but depends only on its corresponding input context.

\paragraph{{\bf Maintaining Holds-Time}} 
Now, the timestamp argument is added to the transform:
\[
\begin{array}{lcl}
q(I,O,H) &\leftarrow&\#r(q,[cons(a)],I,R,H_r), cons(a,R,O,H_a),\\
& &  latest([\#r(q,[cons(a)],I,R,H_r),cons(a,R,O,H_a)],H)\text{.}\\
expect(a,I,I,1)\text{.} & &
\end{array}
\]
The time when $q$ is true (holds-time $H$ of $q$) is derived from the holds-time $H_r$ of its rule name fluent $\#r(q,[consider(a)])$ and $H_a$ of $consider(a)$, via the $latest/2$ reserved predicate. Conceptually, $H$ is determined by which \emph{inertial} fluent in its body holds \emph{latest}. Therefore, the predicate $latest(Body,H)$ does not merely find the maximum $H$ of $H_a$ and $H_r$, but also assures that no fluent in $Body$ was subsequently supervened by its complement at some time up to $H$. The holds-time for $expect(a)$ is set to 1, by convention the initial time when the program is inserted.

Finally, recursion through frame axiom can be avoided by tabling fluents -- in essence, tabling their holds-time -- so it is enough to look-up the time these fluents are true in the table, and pick-up the most recent holds-time. For this purpose, incremental tabling is employed to ensure the consistency of answers in the table due to updates or changes on which the table depends, by incrementally maintaining the table through updates propagation. Similar to \er, the incremental tabling of fluents is achieved via a reserved incrementally tabled predicate $fluent(F,I,O,H)$, defined as follows:
\[\texttt{:- } table\ fluent/4\ as\ incremental\text{.}\]
\[fluent(F,I,O,H) \leftarrow upper(Lim), extend(F,[I,O,H],F'), call(F'), H \leq Lim\text{.}\]
where $extend(F,Args,F')$ extends the arguments of fluent $F$ with those in list $Args$ to obtain $F'$. The definition requires a predefined upper time limit $Lim$, which is used to delimit updates propagation due to potential iterative non-termination propagation, cf. \cite{lpar19-evolpr} for details. Since $fluent(F,I,O,H)$ simply calls fluent $F$ with a given list of context arguments $I$, $O$, and holds-time $H$, calls to fluents in the body of a rule can be recast into calls via reserved predicate $fluent/4$. The above transform finally becomes:
\[\texttt{:- } dynamic\ \#r/5, expect/4\ as\ incremental\text{.}\]
\[
\begin{array}{lcl}
q(I,O,H) &\leftarrow&fluent(\#r(q,[cons(a)]),I,R,H_r),\\
& & cons(a,R,O,H_a),\\
& &  latest([\#r(q,[cons(a)],I,R,H_r),cons(a,R,O,H_a)],H)\text{.}\\
expect(a,I,I,1)\text{.} & &
\end{array}
\]
along with the assertion of rule name fluent $\#r(q,[cons(a)])$ at the initial time 1,

{\centering

 $\#r(q,[cons(a)],I,I,1)$.\\
 }
\noindent
Note that rule name predicate $\#r/5$ and predicate $expect/4$ may be subjected to incremental updates, hence their declaration as dynamic and incremental. On the other hand, predicate $consider/4$ (i.e., $cons/4$ in the example) is not so declared, though it depends (directly or indirectly) on dynamic incremental predicates $expect/4$ and $expect\_not/4$, as we further show in the subsequent section. Thus, there is no need to wrap its call in the body with the reserved predicate $fluent/4$.

\paragraph{{\bf Tabling of Abductive Solutions}}
In the preprocessing, cf. ``Enabling Abducibles'', every abducible $A$ appearing in the body of a rule is substituted with $consider(A)$. Recall the definition of $consider(A)$:

{\centering

$consider(A) \leftarrow expect(A),\ not\ expect\_not(A),\ A$.\\
}
\noindent
After preprocessing, the abducible $A$ thus only appears in the definition of $consider(A)$. Consequently, the transformation that deals with tabling of abductive solutions takes place only in the definition of $consider/1$. Like in \tabdual, we introduce two new predicates for $consider/1$, namely $consider_{ab}/3$ and $consider/4$, where predicate $consider_{ab}/3$ is used to table an abductive solution. We first define $consider_{ab}/3$ ($exp$ is shorthand for $expect$):
\[\hspace{-7mm}\texttt{:- } table\ consider_{ab}/3\ as\ incremental\text{.} \]
\[
\hspace{-7mm}
\begin{array}{lcl}
consider_{ab}(A,E,T) & \leftarrow & timed(A,A_T),\\
& &  fluent(exp(A),[A_T],R,H_1),\\
& & fluent(not\_exp\_not(A),R,E,H_2),\\
& & latest([exp(A,[A_T],R,H_1),not\_exp\_not(A,R,E,H_2)],T)\text{.}
\end{array}
\]
Observe that the tabled abductive solution entry $E$ is derived by relaying the ongoing abductive solution stored in context $R$ from subgoal $fluent(exp(A),[A_T],R,H_1)$ to subgoal $fluent(not\_exp\_not(A),R,E,H_2)$ in the body, given $[A_T]$ as the input abductive context of $exp(A)$. This input context $[A_T]$ comes from the abducible $A$ appearing in the body of $consider(A)$ after it is equipped with $T$, i.e., the time $A$ is abduced; $A_T$ is obtained using predicate $timed(A,A_T)$. Notice that time $T$ is the same time that $consider_{ab}(A)$ is true, which is the latest time between the two fluents, $exp(A)$ and $not\_exp\_not(A)$. Notice also that the subgoal call $not\ expect\_not(A)$ in the original definition becomes a predicate $not\_exp\_not(A)$ in the subgoal call $fluent/4$, in the transform. This predicate is the dual of $exp\_not$ and is obtained by the dual transformation, as explained in the next section. Like $expect/4$, it is subject to updating, and thus, declared as dynamic and incremental too.

Next, we define predicate $consider/4$, which reuses the tabled solution entry $E$ from $consider_{ab}/3$, for a given input context $I$, to obtain a solution in its output context $O$. It is defined as (the holds-time $H$ is just passed from the body to the head):

{\centering

$consider(A,I,O,H) \leftarrow consider_{ab}(A,E,H), produce(O,I,E)$.\\
}
\noindent
The reserved predicate $produce(O,I,E)$ should guarantee that it produces a \emph{consistent} output context $O$ from $I$ and $E$ that encompasses both. For instance, $produce(O,[b_3],[a_1])$ and $produce(O,[a_1,b_3],[a_1])$ both succeed with $O=[a_1,b_3]$, but $produce(O,[not\ a_1],[a_1])$ fails because conjoining $E=[a_1]$ and $I=[not\ a_1]$ results in an inconsistent abductive context $O=[a_1,not\ a_1]$.

\paragraph{{\bf The Dual Program Transformation}}
The different purposes of the dual program transformation in \tabdual and \er, cf. Section \ref{sec:prelim},  are consolidated in the integration. First, the dual predicate $not\_G$ for the negation of goal $G$ in \tabdual and $\til G$ for the negation complement of fluent $G$ in \er are now represented uniquely as $not\_G$, declared dynamic and incremental. Second, the abductive context and holds-time arguments jointly figure in dual predicates, as for the positive transform.

The reader is referred to \cite{tr13-tabdual} for a formal specification and refinement of the dual transformation. We illustrate the transformation for $q/0$ and $expect/1$ of Example \ref{ex:incr-tabling}. With regard to $q$, the transformation will create dual rules for $q$ that falsify $q$ with respect to its only rule,\footnote{In general, if $q$ is defined by $n$ rules, then $not\_q$ is obtained by falsifying each of these $n$ rules, i.e., it is defined as the conjunction of $q^{\ast 1},\dots,q^{\ast n}$ and relays the ongoing abductive solution from $q^{\ast i}$ to $q^{\ast (i+1)}$ via abductive contexts. The holds-time of $not\_q$ is obtained as in the positive transform, i.e., via reserved predicate $latest/2$ from each holds-time of inertial dualized literals in $q^{\ast 1},\dots,q^{\ast n}$.} expressed by predicate $q^{\ast 1}$: 

{\centering
$not\_q(I,O,H)\leftarrow q^{\ast 1}(I,O,H)$.

}
\noindent

Next, predicate $q^{\ast 1}$ is defined by falsifying the body of $q$'s rule in the transform. That is, the rule of $q$ is falsified by alternatively failing one subgoal in its body at a time, i.e. by negating $\#r(q,[cons(a)])$ or, instead, by negating $consider(a)$ and keeping $\#r(q,[cons(a)])$. Therefore, we have:
\[
\begin{array}{lcl}
q^{\ast 1}(I,O,H) & \leftarrow & fluent(not\_\#r(q,[cons(a)]),I,O,H)\text{.}\\
q^{\ast 1}(I,O,H) & \leftarrow & fluent(\#r(q,[cons(a)]),I,R,H_r), not\_consider(a,R,O,H),\\
& & verify\_pos([\#r(q,[cons(a)],I,R,H_r)],H)\text{.}
\end{array}
\]
Observe that in both rules, the holds-time of $q^{\ast 1}$ is determined by the dualized goal in the body, i.e., $fluent(not\_\#r(q,[cons(a)]),I,O,H)$ in case of the first rule, and $not\_consider(a,R,O,H)$ in case of the second. Because the final solution in $O$ is obtained from the intermediate contexts of the preceding positive goals, the reserved predicate $verify\_pos(Pos,H)$ ensures that none of the positive goals in $Pos$ were subsequently supervened by their complements at some time up to $H$.

With regard to $expect/1$, we have the dual rules:

{\centering

$not\_expect(A,I,O,H)\leftarrow expect^{\ast 1}(A,I,O,H)$.\qquad $expect^{\ast 1}(A,I,I,H) \leftarrow A \neq a$.\\
}
\noindent
The uninstantiated holds-time $H$ may get instantiated later, possibly in conjunction with other goals, or if it does not, eventually so by the actual query time. The input context $I$ of $expect^{\ast 1}$  is simply relayed to its output, since $A\neq a$ induces no abduction at all.

Finally, the dual of $consider(A)$ is defined as ($exp$ is shorthand for $expect$):
\[
\hspace{-7mm}
\begin{array}{lcl}
not\_consider(A,I,O,H) &\leftarrow &consider^{\ast 1}(A,I,O,H)\text{.}\\
consider^{\ast 1}(A,I,O,H)&\leftarrow & not\_A(I,O,H)\text{.}\\ 
consider^{\ast 1}(A,I,O,H)&\leftarrow & fluent(not\_exp(A),I,O,H)\text{.}\\
consider^{\ast 1}(A,I,O,H)&\leftarrow & fluent(exp(A),I,R,H_e), fluent(exp\_not(A),R,O,H),\\
 & & verify\_lits([exp(A,I,R,H_e)],H)\text{.}
 \end{array}
\]
In the first rule of $consider^{\ast 1}$, the negation of $A$, i.e. $not\ A$, is abduced by invoking the subgoal $not\_A(I,O,H)$. This subgoal is defined via the transformation of abducibles below (say for $not\_a$):

{\centering
$not\_a(I,O,H) \leftarrow insert(not\ a(H),I,O)$.

}
\noindent
where $insert(A,I,O)$ is a reserved predicate that inserts abducible $A$ into input context $I$, resulting in output context $O$, while also keeping the consistency of the context (like in $produce/3$). Again, the holds-time $H$ may get instantiated later, like in the case of $not\_expect/4$, above.

\paragraph{{\bf The Top-Goal Query}}
As in \er, updates propagation by incremental tabling is query-driven, i.e., the actual query time is used to control updates propagation by first keeping the sequence of updates pending, say in the database, and then only making active, through incremental assertions, those with timestamps up to the actual query time (if they have not yet been so made already by queries of a later timestamp). Given that an upper time limit has been set (cf. $fluent/4$ definition) and that some pending updates may be available, the system is ready for a top-goal query. The query $holds(G,I,O,Qt)$ determines the truth and the abductive solution $O$ of goal $G$ at query time $Qt$, given input context $I$. It is defined as:
\[
\hspace{-7mm}
\begin{array}{lcl}
holds(G,I,O,Qt) & \leftarrow & activate\_pending(Qt),\ compl(G,G'),\\
& & compute(G,I,O,H,Qt,V),\ compute(G',I,O,H',Qt,V'),\\
& & verify\_holds(H,V,H',V')\text{.}
\end{array}
\]
where $activate\_pending(Qt)$ activates all pending updates up to $Qt$ and $compl(G,G')$ obtains the dual complement $G'$ from $G$. The reserved predicate $compute(G,I,O,H,Qt,V)$ returns the highest timestamp $H\leq Qt$ of goal $G$, and its abductive solution $O$, given input context $I$. It additionally returns the truth value $V$ of $G$, obtained through the XSB predicate $call\_tv/2$. This is achieved by $call\_tv(fluent(G,I,O,H),V)$, where $V$ may be instantiated with $true$ or $undefined$.\footnote{Fluents, that are not defined in the program by any rule or fact, have the truth value $undefined$ at the initial time 1. In this case, the content of its input context is simply relayed to its output one. Such fluents inertially remain $undefined$ at query time $Qt$, if they are never updated up to $Qt$.} Finally, the predicate $verify\_holds(H,V,H',V')$ ensures that $H\geq H'$, and determines the truth value of $G$ based on $V$ and $V'$. Note that, when $compute(F,I,O,H,Qt,V)$ fails, by convention it returns $V=false$ with $H=0$ (the output context $O$ is ignored). This is merely for a technical reason, to prevent $compute/6$ failing prematurely before $verify/4$ is called. 

\section{Concluding Remarks}\label{sec:conclude}
\paragraph{{\bf Related Work}}
Abductive logic programming with destructive databases \cite{ks11-alpdd} is a distinct but somewhat similar and complementary to ours. It defines an agent language based on abductive logic programming and relies on the fundamental role of state transition systems in computing, realizing fluent updates by destructive assignment. Their approach differs from ours in that it defines a new language and an operational semantics, rather than taking an existing one. Moreover, it is implemented in LPA Prolog with no underlying tabling mechanisms, whereas in our work both abduction and fluent updates are managed by tabling mechanisms supported by XSB Prolog.

The connection of knowledge updates and abduction is also studied in \cite{si99-updating}, where techniques for updating knowledge bases are introduced and formulated through abduction. On the other hand, the technique we propose pertains to the integration of abduction and logic program updates via tabling, with no focus on formulating updates by means of abduction. Our approach also makes use of abductive contexts, making it suitable for contextual abductive reasoning.

A dynamic abductive logic programming procedure, called LIFF, is introduced in \cite{st06-interleaving}. It allows reasoning in dynamic environments without the need to discard earlier reasoning when changes occur. Though in that work updates are assimilated into abductive logic programs, its emphasis is distinct from ours, as we do not propose a new proof procedure in that respect, but rather an implementation technique using a pre-existing theoretical basis.

Updates propagation has been well studied in the context of deductive databases, e.g., extending the SLDNF procedure for updating knowledge bases while maintaining their consistency, including integrity constraints maintenance \cite{to95-updating}, using abduction for view updating \cite{decker96-view}, as well as fixpoint approaches \cite{behrend11-fixpoint}. Though these methods do not directly deal with tabling mechanisms for the integration of abduction and logic program updates, the approaches proposed in those works seem relevant to ours and some cross-fertilization may lead to gains.

\paragraph{{\bf Conclusion and Future work}}
In this work we have proposed a novel logic programming implementation technique that aims at integrating abduction and logic program updates by means of innovative tabling mechanisms. We have based the present work on our two previously devised techniques, viz., tabled abduction (\tabdual) and query-driven updates propagation by incremental tabling (\er).  The main idea of the integration is to fuse and to mutually benefit from tabling features already employed in each of our previous approaches, and is afforded by a new program transformation synthesis, and library of reserved predicates. The current implementation has simplified the transformation to some extent, e.g., using tries data structure to construct dual rules only as they are needed (like in \tabdual).  Future work consists in perfecting the implementation and conducting experimental evaluation to validate the implementation. We aim at deploying it in an agent life cycle comprising hypothetical reasoning, counterfactual, and moral decision making, which we are currently pursuing.

\paragraph{\textbf{Acknowledgements}}
Ari Saptawijaya acknowledges the support of FCT/MEC Portugal, grant SFRH/BD/72795/2010.

\bibliographystyle{acmtrans}
\bibliography{bibs}

\begin{thebibliography}{}

\bibitem[\protect\citeauthoryear{Alferes, Brogi, Leite, and Pereira}{Alferes
  et~al\mbox{.}}{2002}]{ablp02-evolp}
{\sc Alferes, J.~J.}, {\sc Brogi, A.}, {\sc Leite, J.~A.}, {\sc and} {\sc
  Pereira, L.~M.} 2002.
\newblock Evolving logic programs.
\newblock In {\em JELIA 2002}. LNCS, vol. 2424. Springer, 50--61.

\bibitem[\protect\citeauthoryear{Alferes, Pereira, and Swift}{Alferes
  et~al\mbox{.}}{2004}]{aps04-abdual}
{\sc Alferes, J.~J.}, {\sc Pereira, L.~M.}, {\sc and} {\sc Swift, T.} 2004.
\newblock Abduction in well-founded semantics and generalized stable models via
  tabled dual programs.
\newblock {\em Theory and Practice of Logic Programming\/}~{\em 4,\/}~4,
  383--428.

\bibitem[\protect\citeauthoryear{Behrend}{Behrend}{2011}]{behrend11-fixpoint}
{\sc Behrend, A.} 2011.
\newblock A uniform fixpoint approach to the implementation of inference
  methods for deductive databases.
\newblock In {\em INAP 2011}.

\bibitem[\protect\citeauthoryear{Decker}{Decker}{1996}]{decker96-view}
{\sc Decker, H.} 1996.
\newblock An extension of sld by abduction and integrity maintenance for view
  updating in deductive databases.
\newblock In {\em Procs. of the 1996 Joint International Conference and
  Symposium on Logic Programming}.

\bibitem[\protect\citeauthoryear{Denecker and de~Schreye}{Denecker and
  de~Schreye}{1992}]{ds92-sldnfa}
{\sc Denecker, M.} {\sc and} {\sc de~Schreye, D.} 1992.
\newblock {SLDNFA}: {A}n abductive procedure for normal abductive programs.
\newblock In {\em Procs. of the Joint Intl. Conf. and Symp. on Logic
  Programming}. The MIT Press.

\bibitem[\protect\citeauthoryear{Eiter, Gottlob, and Leone}{Eiter
  et~al\mbox{.}}{1997}]{egl97-abduction}
{\sc Eiter, T.}, {\sc Gottlob, G.}, {\sc and} {\sc Leone, N.} 1997.
\newblock Abduction from logic programs: semantics and complexity.
\newblock {\em Theoretical Computer Science\/}~{\em 189,\/}~1-2, 129--177.

\bibitem[\protect\citeauthoryear{Fung and Kowalski}{Fung and
  Kowalski}{1997}]{fk97-iff}
{\sc Fung, T.~H.} {\sc and} {\sc Kowalski, R.} 1997.
\newblock The {IFF} procedure for abductive logic programming.
\newblock {\em Journal of Logic Programming\/}~{\em 33,\/}~2, 151--165.

\bibitem[\protect\citeauthoryear{Inoue and Sakama}{Inoue and
  Sakama}{1996}]{is96-fixpoint}
{\sc Inoue, K.} {\sc and} {\sc Sakama, C.} 1996.
\newblock A fixpoint characterization of abductive logic programs.
\newblock {\em J. of Logic Programming\/}~{\em 27,\/}~2, 107--136.

\bibitem[\protect\citeauthoryear{Kakas, Kowalski, and Toni}{Kakas
  et~al\mbox{.}}{1998}]{kkt98-alp}
{\sc Kakas, A.}, {\sc Kowalski, R.}, {\sc and} {\sc Toni, F.} 1998.
\newblock The role of abduction in logic programming.
\newblock In {\em Handbook of Logic in Artificial Intelligence and Logic
  Programming}, {D.~Gabbay}, {C.~Hogger}, {and} {J.~Robinson}, Eds. Vol.~5.
  Oxford U. P.

\bibitem[\protect\citeauthoryear{Kowalski and Sadri}{Kowalski and
  Sadri}{2011}]{ks11-alpdd}
{\sc Kowalski, R.} {\sc and} {\sc Sadri, F.} 2011.
\newblock Abductive logic programming agents with destructive databases.
\newblock {\em Annals of Mathematics and Artificial Intelligence\/}~{\em
  62,\/}~1, 129--158.

\bibitem[\protect\citeauthoryear{Pereira, Dell'Acqua, Pinto, and Lopes}{Pereira
  et~al\mbox{.}}{2013}]{pdpl13-inspecting}
{\sc Pereira, L.~M.}, {\sc Dell'Acqua, P.}, {\sc Pinto, A.~M.}, {\sc and} {\sc
  Lopes, G.} 2013.
\newblock Inspecting and preferring abductive models.
\newblock In {\em The Handbook on Reasoning-Based Intelligent Systems},
  {K.~Nakamatsu} {and} {L.~C. Jain}, Eds. World Scientific Publishers,
  243--274.

\bibitem[\protect\citeauthoryear{Pereira, Dietz, and H\"olldobler}{Pereira
  et~al\mbox{.}}{2014}]{pdh14-contextual}
{\sc Pereira, L.~M.}, {\sc Dietz, E.-A.}, {\sc and} {\sc H\"olldobler, S.}
  2014.
\newblock Contextual abductive reasoning with side-effects.
\newblock In {\em ICLP 2014}.

\bibitem[\protect\citeauthoryear{Poole}{Poole}{1988}]{poole88-default}
{\sc Poole, D.~L.} 1988.
\newblock A logical framework for default reasoning.
\newblock {\em Artificial Intelligence\/}~{\em 36,\/}~1, 27--47.

\bibitem[\protect\citeauthoryear{Sadri and Toni}{Sadri and
  Toni}{2006}]{st06-interleaving}
{\sc Sadri, F.} {\sc and} {\sc Toni, F.} 2006.
\newblock Interleaving belief updating and reasoning in abductive logic
  programming.
\newblock In {\em ECAI 2006}. Frontiers of Artificial Intelligence and
  Applications (FAIA), vol. 141. IOS Press, 442--446.

\bibitem[\protect\citeauthoryear{Saha}{Saha}{2006}]{saha06-incremental}
{\sc Saha, D.} 2006.
\newblock Incremental evaluation of tabled logic programs.
\newblock Ph.D. thesis, SUNY Stony Brook.

\bibitem[\protect\citeauthoryear{Sakama and Inoue}{Sakama and
  Inoue}{1999}]{si99-updating}
{\sc Sakama, C.} {\sc and} {\sc Inoue, K.} 1999.
\newblock Updating extended logic programs through abduction.
\newblock In {\em LPNMR 1999}. LNAI, vol. 1730. Springer, 147--161.

\bibitem[\protect\citeauthoryear{Saptawijaya and Pereira}{Saptawijaya and
  Pereira}{2013a}]{lpar19-evolpr}
{\sc Saptawijaya, A.} {\sc and} {\sc Pereira, L.~M.} 2013a.
\newblock Incremental tabling for query-driven propagation of logic program
  updates.
\newblock In {\em LPAR-19}. LNCS ARCoSS, vol. 8312. Springer, 694--709.

\bibitem[\protect\citeauthoryear{Saptawijaya and Pereira}{Saptawijaya and
  Pereira}{2013b}]{lpnmr13-evolpr}
{\sc Saptawijaya, A.} {\sc and} {\sc Pereira, L.~M.} 2013b.
\newblock Program updating by incremental and answer subsumption tabling.
\newblock In {\em LPNMR 2013}. LNCS, vol. 8148. Springer, 479--484.

\bibitem[\protect\citeauthoryear{Saptawijaya and Pereira}{Saptawijaya and
  Pereira}{2013c}]{tr13-tabdual}
{\sc Saptawijaya, A.} {\sc and} {\sc Pereira, L.~M.} 2013c.
\newblock Tabled abduction in logic programs.
\newblock Tech. rep., CENTRIA, Departamento de Inform{\'a}tica, Faculdade de
  Ci{\^e}ncias e Tecnologia, Universidade Nova de Lisboa. Available at
  \url{http://centria.di.fct.unl.pt/~lmp/publications/online-papers/tabdual_lp.pdf}.

\bibitem[\protect\citeauthoryear{Saptawijaya and Pereira}{Saptawijaya and
  Pereira}{2013d}]{iclp13-tabdual}
{\sc Saptawijaya, A.} {\sc and} {\sc Pereira, L.~M.} 2013d.
\newblock Tabled abduction in logic programs ({T}echnical {C}ommunication of
  {ICLP 2013}).
\newblock {\em Theory and Practice of Logic Programming, Online
  Supplement\/}~{\em 13,\/}~4-5.

\bibitem[\protect\citeauthoryear{Saptawijaya and Pereira}{Saptawijaya and
  Pereira}{2014}]{padl14-moral}
{\sc Saptawijaya, A.} {\sc and} {\sc Pereira, L.~M.} 2014.
\newblock Towards modeling morality computationally with logic programming.
\newblock In {\em PADL 2014}. LNCS, vol. 8324. Springer, 104--119.

\bibitem[\protect\citeauthoryear{Satoh and Iwayama}{Satoh and
  Iwayama}{2000}]{si00-abductionTMS}
{\sc Satoh, K.} {\sc and} {\sc Iwayama, N.} 2000.
\newblock Computing abduction by using {TMS} and top-down expectation.
\newblock {\em Journal of Logic Programming\/}~{\em 44,\/}~1-3, 179--206.

\bibitem[\protect\citeauthoryear{Swift and Warren}{Swift and
  Warren}{2010}]{sw10-answer}
{\sc Swift, T.} {\sc and} {\sc Warren, D.~S.} 2010.
\newblock Tabling with answer subsumption: Implementation, applications and
  performance.
\newblock In {\em JELIA 2010}. LNCS, vol. 6341. Springer, 300--312.

\bibitem[\protect\citeauthoryear{Swift and Warren}{Swift and
  Warren}{2012}]{sw12-xsb}
{\sc Swift, T.} {\sc and} {\sc Warren, D.~S.} 2012.
\newblock {XSB}: Extending {P}rolog with tabled logic programming.
\newblock {\em Theory and Practice of Logic Programming\/}~{\em 12,\/}~1-2,
  157--187.

\bibitem[\protect\citeauthoryear{Teniente and Oliv\'e}{Teniente and
  Oliv\'e}{1995}]{to95-updating}
{\sc Teniente, E.} {\sc and} {\sc Oliv\'e, A.} 1995.
\newblock Updating knowledge bases while maintaining their consistency.
\newblock {\em The VLDB Journal\/}~{\em 4,\/}~2, 193--241.

\end{thebibliography}

\label{lastpage}
\end{document}